\pdfoutput=1
\documentclass[letterpaper]{article} 
\usepackage[preprint]{aaai2027}  
\usepackage[hyphens]{url}  
\usepackage{graphicx} 
\usepackage{natbib}  
\usepackage{caption} 
\usepackage{algorithm}
\usepackage{algorithmic}
\usepackage{booktabs}
\usepackage{amsmath}
\usepackage{amssymb}
\usepackage{xcolor}
\usepackage{array}
\usepackage[most]{tcolorbox}

\definecolor{mblue}{HTML}{2A78D6}
\definecolor{mbluedk}{HTML}{1C5CAB}
\definecolor{mbluebg}{HTML}{EAF2FC}
\definecolor{sorange}{HTML}{EB6834}
\definecolor{sorangedk}{HTML}{B84A1D}
\definecolor{sorangebg}{HTML}{FDF0E9}
\definecolor{agreen}{HTML}{1E7A46}
\definecolor{agreenbg}{HTML}{ECF7F0}
\definecolor{cardgray}{HTML}{5C6670}
\definecolor{cardgraybg}{HTML}{F4F5F6}
\definecolor{inkmut}{HTML}{52514E}

\usepackage{hyperref}
\hypersetup{
  colorlinks=true,
  linkcolor=mbluedk,   
  citecolor=mbluedk,   
  urlcolor=mbluedk,
  linktoc=all
}

\newcommand{\algkey}[1]{{\setlength{\fboxsep}{1.5pt}\colorbox{mbluebg}{#1}}}

\newcommand{\cardsub}[1]{\hfill{\normalfont\footnotesize\color{inkmut}#1}}
\newtcolorbox{taskbox}[1]{enhanced, colback=cardgraybg, colframe=cardgray,
  boxrule=0.8pt, arc=2mm, left=6pt, right=6pt, top=4pt, bottom=4pt,
  fonttitle=\small\bfseries, coltitle=black, colbacktitle=cardgraybg,
  title=#1, toptitle=2.5pt, bottomtitle=0pt, before skip=6pt, after skip=6pt}

\title{Who Grades the Grader?\\Co-Evolving Evaluation Metrics and Skills for Self-Improving LLM Agents}

\author{
    Xing Zhang\textsuperscript{\rm 1},
    Guanghui Wang\textsuperscript{\rm 1},
    Yanwei Cui\textsuperscript{\rm 1},
    Ziyuan Li\textsuperscript{\rm 2},
    Wei Qiu\textsuperscript{\rm 2},
    Bing Zhu\textsuperscript{\rm 2},
    Peiyang He\textsuperscript{\rm 1}\thanks{Corresponding author: peiyan@amazon.com}
}
\affiliations{
    \textsuperscript{\rm 1}AWS Generative AI Innovation Center\\
    \textsuperscript{\rm 2}HSBC Holdings Plc., HSBC Technology Center, China
}

\begin{document}

\maketitle

\begin{abstract}
Self-evolving agent systems create, revise, and retire their own skills, but every such loop assumes a reliable evaluation metric already exists. In many real applications none does. We show the metric itself can be the evolving object: our loop searches compositions of small typed drawback detectors under a full evolutionary lifecycle, selecting for agreement with a ten-item anchored reference set and regularizing by consensus over unlabeled outputs. What evolves is the function that grades one output, never the fixed task sets it is scored on, and what comes out is an inspectable expression rather than an opaque judge. It is also valid: on code generation it gains 0.21 agreement with hidden ground truth on a locked set that metric selection never reads (paired $p=0.014$), beating the bare LLM judge it contains. Validity is where safety lives: removing the anchor guards collapses the metric into a vacuous always-pass detector while removing the detector lifecycle does not, inverting the lesson from skill evolution. That collapse warns this line of work that downstream task score cannot validate a self-evolved evaluator, since the collapsed metric trains skills just as well. Task score answers only sufficiency, and an evolved metric suffices: \emph{Double Ratchet}, co-evolving the metric with a lifecycle-managed skill loop, retains 88--110\% of the lift ground truth or a hand-written rubric buys, across MBPP+, Spider~2.0-Snow, and report generation. When evolved skills gamed the report rubric, an independent judge caught it and one added detector repaired it.
\end{abstract}

\section{Introduction}

Agent self-evolution is now a standard recipe: a frozen language model is wrapped in a loop that writes skills, measures their contribution, and retires the ones that hurt \citep{zhang2026ratchet,wang2023voyager,shinn2023reflexion}. Every round consumes one expensive ingredient: an evaluation signal deciding whether an attempt passed or failed. Unit tests supply it free on code benchmarks; in deployed systems it is the bottleneck. Industry products fill the gap by hand, one rubric or demonstration at a time: Claude's managed agents have developers write outcome rubrics for a separate grader \citep{anthropic2026outcomes}, and Codex's Record \& Replay has a human demonstrate each workflow once \citep{openai2026recordreplay}. This paper automates that step.\footnote{Code: \url{https://github.com/amazon-science/Self-Evolving-Agents-Double-Ratchet}.}

The question is direct: if skills can be evolved, why not the metric? To our knowledge this is the first system to treat the evaluation metric as a first-class evolving object under a full evolutionary lifecycle in the \emph{sparse-anchor} regime: no evaluator exists to score the search, only ten anchored items with golden references. Sparse is not unsupervised, and those ten items are what the method turns on. Existing evolutionary search over objectives is scored by a downstream metric assumed valid, and learned evaluators assume exactly the ground truth we remove; here selection sees only agreement with a ten-item anchor plus consensus over unlabeled outputs, and what comes out is an \emph{inspectable expression} rather than an opaque scalar.

Two clarifications determine how every result should be read. First, \textbf{the metric is the function that scores one solver output}, and it is the only thing that evolves: not the benchmark, whose tasks and references are fixed before any loop starts, so no result here can come from an easier test. Second, \textbf{an evolved metric is judged on validity, not on the task score it produces}: agreement with a locked reference is the claim, while how well skills trained on it perform is weaker evidence, since a metric that passes everything still leaves useful practice behind. That second point is not a caveat we inherited but a result we measured, and it is strong enough to constrain how this literature should run its experiments at all.

\begin{figure*}[t]
\centering
\includegraphics[width=\textwidth]{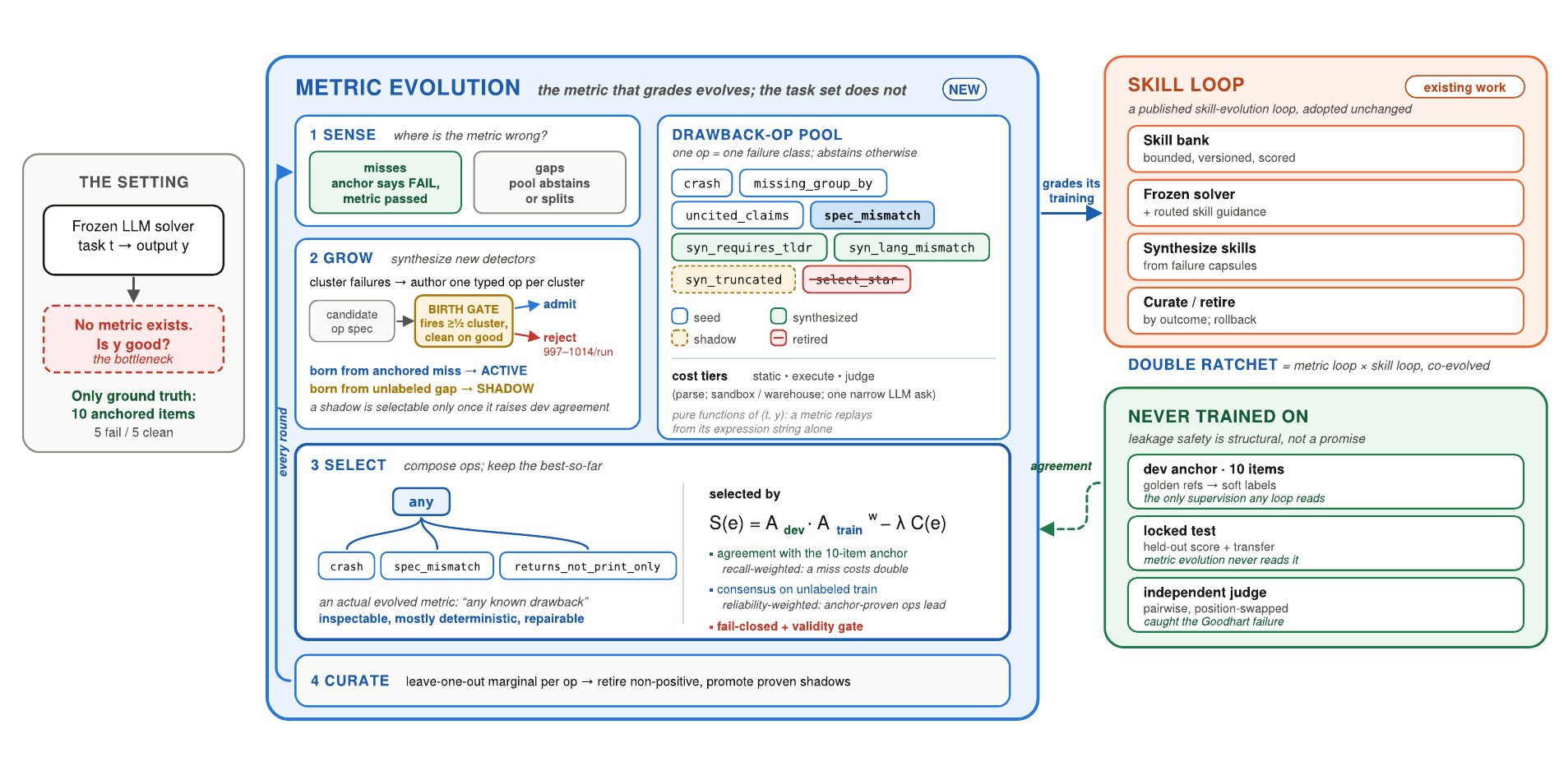}
\caption{\textbf{The evaluation metric is the evolving object; the task sets are fixed and never altered.} \emph{Center:} from a frozen solver, no metric, and ten anchored items (left), the metric loop synthesizes typed drawback detectors behind a birth gate, composes them into an expression selected by anchor agreement and consensus, and curates the pool by marginal contribution. What comes out is a readable expression, not a learned scalar. \emph{Right:} the skill loop is Ratchet \citep{zhang2026ratchet}, adopted unchanged, and the evolved metric grades its training, giving \emph{Double Ratchet}. \emph{Bottom right:} the splits metric evolution never trains on.}
\label{fig:hero}
\end{figure*}

Concretely (Figure~\ref{fig:hero}), the metric is an expression tree over a pool of small drawback detectors, each checking one failure class, synthesized from clustered failures, gated at birth, scored by marginal contribution, and retired when useless. One stance underpins the design: we rarely know what good is, but given an output we can usually find drawbacks, so a clean verdict means no known drawback was found, not certified correctness. That is what makes bootstrap possible, and mostly deterministic detectors, which fail differently from the LLM being graded, resist the shared-blind-spot collusion a bare LLM judge invites. Because a metric grading the loop that produced it is a Goodhart hazard \citep{goodhart1984problems,gao2023scaling}, anchor discipline is strict: the tiny dev set is the only supervised signal, and no part of metric evolution reads the locked set.

The hypothesis comes from the skill side: Ratchet \citep{zhang2026ratchet,zhang2026librarydrift} found unmanaged skill libraries drift and that lifecycle management is what makes skill evolution work at all. Do evaluators need the same discipline? Two ablations judged on \emph{validity} locate which ingredient carries the load, and the answer is not the one the skill literature predicts. Of our three task families, MBPP+ and Spider~2.0-Snow carry exact ground truth, so validity and sufficiency are both measurable; report generation is the intended target, having no golden metric at all, and there we add the layer a deployment would keep, an independent stronger judge comparing final outputs pairwise against the pre-evolution baseline. That audit caught a real Goodhart failure, and auditing the judge exposed a second one. Verifier-free self-improvement is feasible, but only inside an architecture that expects its own metric to be gamed and its own judge miscalibrated, and can catch both.

Our contributions are as follows.
\begin{itemize}
\item \textbf{Metric evolution.} We formulate the grading function, rather than the task set, as the evolving object: anchored search over compositions of atomic drawback detectors under a full lifecycle of synthesis, birth gating, marginal-contribution scoring, and retirement.
\item \textbf{An objective that works without a downstream metric.} Recall-weighted tiny-anchor agreement paired with reliability-weighted consensus over unlabeled outputs, so consensus regularizes transfer without ever selecting a metric on its own.
\item \textbf{Where safety lives, and why task score cannot find it.} Anchor discipline, not the pool lifecycle, is load-bearing for evaluators, inverting the skill-evolution lesson; and since the collapsed metric matches the working one on task score, validating a self-evolved evaluator by downstream performance is unsound, a caution for co-evolved-judge designs generally.
\item \textbf{Double Ratchet.} Co-evolution of the metric with a skill loop, with evidence that an evolved metric suffices in place of ground truth inside the loop, plus an end-to-end Goodhart catch, its repair, and a judge audit.
\end{itemize}

\section{Related Work}

Three lines of work meet here, and in each the metric is what our method does not get for free.

\paragraph{Self-evolving agents assume the metric.}
These agents grow skill libraries, revise from failure traces, and bank reusable experience \citep{wang2023voyager,shinn2023reflexion,madaan2023selfrefine,zhao2024expel,chen2024automanual,wang2024awm,park2023generative}, a landscape two surveys map \citep{zhang2025surveyselfevolving,wang2024survey}, and a recent wave evolves and benchmarks such libraries \citep{wu2025evolver,yang2026autoskill,xia2026skillrl,alibaba2026trace2skill,zhang2026ecs,wang2026strategygenes,li2026skillsbench}. Ratchet \citep{zhang2026ratchet,zhang2026librarydrift} showed unmanaged libraries drift and that a minimal lifecycle of retirement, bounded capacity, and meta-skill guidance restores the gains; that published loop is adopted unchanged here as the skill-side component, not a contribution of this work. Others search agent code itself, in the tradition of neural architecture search \citep{hu2025automated,elsken2019nas}, or optimize prompts and pipelines \citep{khattab2024dspy,yang2023opro,yuksekgonul2024textgrad,fernando2024promptbreeder}. All optimize \emph{against} a given evaluation signal. We evolve that signal, and ask which part of the discipline carries the safety load once the evaluator moves.

\paragraph{Searched objectives are scored by a metric that exists.}
CSE-Autoloss evolves losses, AutoML-Zero whole learning algorithms, Pruner-Zero pruning metrics, AlphaEvolve program code against supplied evaluators \citep{liu2021cseautoloss,real2020automlzero,dong2024prunerzero,novikov2025alphaevolve}. The search space is analogous, primitives composed into a program, but they evolve the artifact being graded and fitness differs in kind: each is scored by a metric taken as valid, AlphaEvolve stating outright that progress must be machine-gradable, the precondition our regime lacks. This paper supplies that missing foothold: anchored agreement plus consensus in place of a downstream score. The distinction is not cosmetic. When fitness is a trusted external number, a degenerate candidate is caught by construction, since it simply scores badly; when fitness is assembled from the candidate's own agreement with a ten-item anchor and with its peers, degeneracy has to be excluded by explicit guards, which is why the guards and not the lifecycle turn out to be load-bearing here.

\paragraph{Learned graders trade the assumption for opacity.}
Reward design is delicate even in classical RL \citep{ng1999pbrs}; learned reward models are over-optimized by the policies they train, and misspecification produces systematic gaming \citep{ouyang2022rlhf,gao2023scaling,pan2022effects,skalse2022defining,amodei2016concrete}, while AI-feedback and self-rewarding setups let the model judge itself \citep{bai2022constitutional,yuan2024selfrewarding}. We share the hazard but not the structure: our grader is a composition of typed detectors optimized to predict an anchor it never trains on, so it can be read and repaired by hand, and it is never a reward the agent optimizes directly.

\paragraph{LLM judges, and auditing them.}
LLM judges are standard evaluation infrastructure and widely audited for position, verbosity, and self-preference bias \citep{zheng2023judging,dubois2024alpacafarm,gu2024surveyllmjudge,chen2024alignment}. We confine LLM judgment to three narrow guarded roles: soft labels on the ten dev items, one semantic-mismatch detector class, and a position-debiased pairwise audit outside all loops. Agent-as-judge co-evolution \citep{zhuge2024agentjudge} is closest in spirit and validates the judge through the system it grades, which our ablations show a vacuous grader also passes. Our own results add a caution to this literature that we could not have anticipated: the audit needs to know the task's output contract, or it will score a required format as a defect (\S\hyperref[sec:goodhart]{The Goodhart Episode}).

\paragraph{Reference-free evaluators for our task families.}
The evaluators nearest our tasks fix in advance what we synthesize, gate, and retire. CodeScore \citep{dong2024codescore} predicts functional correctness from the execution-labeled supervision we take to be missing, and SQLHD \citep{yang2025sqlhd} needs no ground truth but fixes a hand-authored family of metamorphic relations \citep{chen1998metamorphic}. Either could be registered as one typed op in our pool and would then compete for selection on the anchor like any other, so we read them as components rather than competitors. Agentic text-to-SQL pipelines \citep{deng2025reforce} improve the solver, an orthogonal axis: a better solver changes which failures remain, which is precisely the variable our detectability analysis isolates.

\paragraph{Benchmarks.}
MBPP+ supplies unit-tested Python tasks \citep{austin2021mbpp,liu2023evalplus}, in the execution-graded tradition of HumanEval \citep{chen2021codex}, and Spider~2.0 supplies enterprise text-to-SQL over real warehouses with execution-based grading \citep{lei2025spider2}. We use both as hard anchors for certifying an evolved metric rather than as leaderboard entries: our solver is a single LLM call with no tools or iteration, so absolute pass rates are far below what agentic systems report and are not meant to be compared with them.

\section{Problem Setup}

Let a task family come with tasks $t$ and a frozen solver mapping $t$ to an output $y$. A metric $M$ maps $(t, y)$ to a verdict in $\{\textsc{pass}, \textsc{fail}, \textsc{abstain}\}$. It is $M$ that we search over, with the tasks and references held fixed. Three disjoint splits play the usual roles with one twist each.

\begin{itemize}
\item \textbf{train}, large and unlabeled: solver outputs here expose where candidate metrics disagree or abstain, and in co-evolution the evolved metric grades these tasks for the skill loop. No reference of any kind is attached to a train task.
\item \textbf{dev}, tiny and anchored, ten items throughout: each carries a golden reference against which a teacher emits soft pass/fail labels, the only supervised signal any loop reads. Those references are not themselves a metric, grading only the ten tasks that carry them while the evolved metric must grade arbitrary outputs on tasks that have none.
\item \textbf{test}, locked against the metric: it holds the strongest available reference (unit tests, official execution comparison, the report rubric), and no part of metric evolution reads it, so metric-validity numbers are fully out of sample.
\end{itemize}

The skill loop is the one component that touches the locked set, for held-out scoring and rollback, as published \citep{zhang2026ratchet} and kept unchanged because a deployed agent rolls back against the cases it is judged on. The guarantee is thus asymmetric, and it is worth being explicit about which claim rests on which split. Metric-validity numbers are strictly out of sample: nothing in synthesis, gating, selection, or retirement ever reads the locked set, so an agreement figure there is an honest generalization estimate. Skill-side task scores are not out of sample in the same sense: they are the pass rate of a loop that rolls back on that set. They remain comparable across arms because every arm rolls back identically against the same set, which is what the head-to-head comparison uses them for, and they are never used to certify a metric. Dev is deliberately small because sparse anchoring is realistic: fresh labels in deployed systems arrive as a trickle of demos and incident reviews, not as a labeled corpus.

\subsection{Metrics as Compositions of Drawback Detectors}

An \emph{op} is an atomic drawback detector $o(t, y, c) \rightarrow \{\textsc{drawback}, \textsc{clean}, \textsc{abstain}\}$ with $c$ a shared context, checking exactly one failure class and abstaining otherwise. A task family starts from a handful covering its obvious failure modes, cheap to write by hand or have an LLM draft from the spec, and the lifecycle grows the pool from there. Ops come in three cost tiers: static ops parse the artifact, execution ops run it (sandbox for code, live warehouse for SQL), and judge ops ask an LLM one narrow question. A fixed non-evolving \emph{root} per family validates basic structure (code parses, a report has a title) and caches the parsed artifact the other ops read.

A \emph{metric expression} composes op verdicts with logical operators. Writing $v(e) \in \{1, 0, \bot\}$ for the verdict of expression $e$ on $(t, y)$, where $1$ denotes \textsc{drawback}, $0$ denotes \textsc{clean}, and $\bot$ denotes \textsc{abstain}, the grammar is
\begin{equation}
\label{eq:grammar}
\begin{aligned}
e \mathrel{::=} {}& o
  \;\;\big|\;\; \textstyle\bigvee_{i \leq k} e_i
  \;\;\big|\;\; \textstyle\bigwedge_{i \leq k} e_i
  \;\;\big|\;\; \neg e \\[2pt]
\big|\;\; & \textstyle\big[\sum_{i \leq k} v(e_i) \geq K\big]
  \;\;\big|\;\; \textstyle\big[\sum_{i \leq k} w_i\, v(e_i) \geq \theta\big],
\end{aligned}
\end{equation}
where $|$ separates the alternative node forms rather than denoting disjunction: a node is a leaf op, a disjunction (a drawback if any child finds one), a conjunction (only if all children agree), a negation, an unweighted $K$-of-$k$ vote, or a weighted vote against threshold $\theta$, so applied recursively a metric is a tree of such nodes. Abstaining children are excluded from their combinator, and a node whose children all abstain abstains itself. The output passes when the root finds no drawback, the final verdict being the fixed root conjoined with the evolved expression. Ops are pure functions of $(t, y)$, so verdicts are cached, evaluation is a set of table lookups, and a metric is reproducible from its expression string plus the registered op pool. Negation and weighted votes stay available throughout, but every metric selected in our runs uses only disjunction, conjunction, and votes.

Keeping the metric mostly deterministic is a safety choice rather than an efficiency one. A bare LLM judge shares failure modes with the LLM solver it grades, so the two can drift into agreement before any optimization pressure exists, and the agreement is invisible from the outside because there is nothing to read. A composition of typed detectors fails differently, one leaf at a time, and each leaf names the failure class it fired on. That property is what turned a rubric-gaming episode into a one-line diagnosis later in this paper.

\subsection{The Metric Loop}

Algorithm~\ref{alg:loop} sketches one round. Two signals feed synthesis: \emph{misses}, dev items the current metric passes although their soft label is fail, and \emph{gaps}, train outputs on which the op pool abstains entirely or splits. Both are clustered by failure class under a closed task-specific taxonomy, and a synthesizer authors a typed op spec for any recurring cluster. A new op must pass a birth gate: fire on at least half its cluster and stay clean on known-good outputs. Ops born from anchored misses enter \emph{active} (selectable), their birth evidence already tying them to the anchor; ops born only from unlabeled gaps enter as \emph{shadows}, which record verdicts every round but cannot be selected, since nothing yet shows they track quality rather than consensus noise. A shadow is promoted once its verdicts raise dev agreement.

Candidates come from an LLM composer that sees the op catalog, the current gaps, and an elite history of the best expressions with their shortcomings, plus mutations and crossovers of those elites. Selection keeps the best-so-far candidate by
\begin{equation}
\begin{split}
S(e) \;=\; A_{\mathrm{dev}}(e) \cdot A_{\mathrm{train}}(e)^{w} \;-\; \lambda\, C(e),
\end{split}
\label{eq:score}
\end{equation}
where $A_{\mathrm{dev}}(e)$ is $e$'s agreement with the dev soft labels, $A_{\mathrm{train}}(e)$ its agreement with the consensus verdict of the opining pool ops on unlabeled train outputs, $C(e)$ expression size, and $w, \lambda$ small constants. Consensus is the regularizer: a metric memorizing ten dev items but behaving erratically on the broad distribution loses, while a degenerate always-fail metric loses on dev. Two guards apply: selection fails closed (a candidate with no usable dev opinion is unselectable), and a validity gate drops candidates that pass, fail, or abstain on everything. Per-op fitness is the leave-one-out marginal of the incumbent expression, and non-positive ops retire after a grace period. Every constant, including the exponent $w$, the complexity charges, and the grace length, is listed in \hyperref[app:lang]{Appendix~C}.

Both agreement terms are weighted, each weighting a safety choice rather than a tuning knob. \emph{Dev agreement is recall-weighted}: $A_{\mathrm{dev}}$ is the weighted mean of the two per-class recalls, $(w_{\textsc{f}} r_{\textsc{fail}} + w_{\textsc{p}} r_{\textsc{pass}})/(w_{\textsc{f}} + w_{\textsc{p}})$ with $w_{\textsc{f}} = 2$, $w_{\textsc{p}} = 1$, over the items the metric opines on, so a missed drawback costs twice a false alarm and no candidate reaches $A_{\mathrm{dev}} = 1$ by acing the easier class alone. \emph{Consensus is reliability-weighted}: each opining op's vote is scaled by $1 + \kappa \max(0, m_o)$ for $m_o$ that op's anchored leave-one-out marginal, so anchor-proven ops lead and unproven ops vote near-neutrally. Otherwise a bloc of correlated detectors, cheap to synthesize and liable to share a blind spot, would define the very pseudo-label the regularizer rewards agreeing with. $A_{\mathrm{train}}$ is thus a transfer regularizer pulled toward the anchor, not a popularity vote, and entering multiplicatively against a fail-closed $A_{\mathrm{dev}}$ it never selects a metric alone.

\begin{algorithm}[t]
\caption{One round of metric evolution. Only the audit step reads the locked test set, and it never feeds back. Constants and the full lifecycle are in \hyperref[app:lang]{Appendix~C}.}
\label{alg:loop}
\begin{algorithmic}[1]
\STATE {\color{mbluedk}\textbf{Sense}} \ {\footnotesize\color{inkmut}\emph{(where is the metric wrong?)}}
\STATE \quad misses $\gets$ dev items the incumbent passes but the soft label fails
\STATE \quad gaps $\gets$ train outputs where the pool abstains or splits
\STATE {\color{mbluedk}\textbf{Grow}} \ {\footnotesize\color{inkmut}\emph{(expand the op pool)}}
\STATE \quad cluster misses $\cup$ gaps by failure class; author one typed op per recurring cluster
\STATE \quad birth gate: fires on $\geq$ half its cluster and clean on known-good $\Rightarrow$ admit (\algkey{active if anchored, else shadow})
\STATE {\color{mbluedk}\textbf{Select}} \ {\footnotesize\color{inkmut}\emph{(recombine, keep the best)}}
\STATE \quad candidates $\gets$ LLM compositions over the pool, plus \algkey{mutations and crossovers of elites}
\STATE \quad keep $\arg\max_e S(e)$ over candidates and the incumbent, behind the validity gate (Eq.~\ref{eq:score})
\STATE {\color{mbluedk}\textbf{Curate}} \ {\footnotesize\color{inkmut}\emph{(merit-based lifecycle)}}
\STATE \quad retire ops with non-positive \algkey{leave-one-out marginal}; promote shadow ops that raise dev agreement
\STATE {\color{agreen}\textbf{Audit}} \ {\footnotesize\color{inkmut}\emph{(measurement only, never a training signal)}}
\STATE \quad report agreement against the locked test set
\end{algorithmic}
\end{algorithm}

\subsection{Co-Evolution with the Skill Loop: Double Ratchet}

The skill loop adopted here \citep{zhang2026ratchet} needs one thing our setting does not supply: a grader for its training tasks. Co-evolution hands it the current best evolved metric, giving \emph{Double Ratchet} (Figure~\ref{fig:hero}). The loops alternate under a fixed front-loaded curriculum, metric phases longest early when coverage is worst: the evolved metric grades train attempts, and failures become capsules whose error text feeds skill synthesis. Held-out evaluation and the rollback anchor stay pinned to the locked test set, which the evolved metric never touches, so a corrupted metric slows skill learning without corrupting the measurement. The comparison is the reference skill loop, identical machinery handed the ground-truth metric (or best available rubric) for free.

\paragraph{Two ablations, one per safety ingredient.}
The \textbf{naive} arm disables the anchor guards (fail-closed anchoring, the validity gate, and skill-loop rollback) with the detector lifecycle fully on, the collusion-prone configuration unanchored agent-judge co-evolution implicitly runs. The \textbf{no-lifecycle} arm does the opposite, keeping the guards but disabling the birth gate and merit-based retirement so every synthesized detector enters active and nothing retires: the metric-side analog of an unmanaged skill library. The two arms are deliberately not a factorial. Each isolates one ingredient against an otherwise complete system, which is what a claim about which ingredient is load-bearing needs; a fully crossed design would add an arm with neither guard nor lifecycle whose collapse is already implied by the naive arm.

\paragraph{Independent final judge.}
Where no accurate metric exists, rubric scores are references and not truth, so a final audit sits outside all loops: a stronger LLM compares each final output pairwise against the pre-evolution baseline, judged twice with positions swapped, a win counting only when both orders agree. It mimics the human acceptance step a deployment keeps, and exposed a proxy-gaming failure the loops could not see. The judge is an audit and never a training signal: no gradient, capsule, hint, or selection decision anywhere in the system reads it.

\section{Experimental Setup}

\paragraph{Task families.}
\textbf{MBPP+} \citep{austin2021mbpp,liu2023evalplus}, Python function synthesis with hidden unit tests as the hard anchor, and \textbf{Spider~2.0-Snow} \citep{lei2025spider2}, enterprise text-to-SQL over real Snowflake warehouses anchored by official execution-result comparison, are the validation tasks: their exact ground truth makes both metric validity and the reference-loop match measurable. \textbf{Report generation} is the target task, a deployment-style workload in which a pipeline writes analyst-report sections from audited evidence cards. No golden metric exists for it: the available rubric score (RAQS) plus golden demonstrations is a partial reference signal, not ground truth. Task cards are in \hyperref[app:tasks]{Appendix~A}; tables abbreviate the tasks as MBPP+, Spider, and Report.

\paragraph{Hard-subset construction.}
For MBPP+ and Spider~2.0 we screen every task with five frozen-solver samples, drop those passed on every sample (no lift room), and stratify the rest by screen pass rate. Splits are 60/10/40 for MBPP+, 59/10/40 for Spider~2.0, and 73/10/48 for reports, whose tasks derive from golden demonstrations and need no screening. Dev always holds ten items, five clear failures and five clean positives. The solver is a single direct LLM call with no tools, so absolute numbers are not comparable to leaderboard agents.

\paragraph{Roles and models.}
All experiments use Claude models: every in-loop role (solver, skill synthesizers, teachers, judge ops) is Claude Opus~4.7, the sole exception being the report task's final judge, the stronger Claude Opus~4.8 kept outside all loops. A same-family teacher is not a circular self-grader: it compares candidate outputs against golden references on dev rather than judging unaided, and never sees the locked anchor. Skill loops run 100 rounds; co-evolution interleaves metric phases of 15/8/5/2 rounds between four 25-round skill phases, matching the reference loop's skill budget. Every role is bound per call from a role-to-model table, so repeating any experiment under a different or mixed model assignment is a configuration change rather than a code change; we hold that table uniform here to keep the comparisons clean (\hyperref[app:repro]{Appendix~H}).

\section{Results}

The findings run along two axes, in a deliberate order: first \emph{metric validity}, agreement with a locked reference no part of metric evolution reads, which is the claim and which the guard ablations then localize; then \emph{task score}, which we show cannot certify a metric and so use only to ask whether an evolved one suffices in place of ground truth. Each run uses three seeds, reported as mean and sample standard deviation, with loop comparisons adding the seed-level difference and a 95\% bootstrap interval. Three seeds and held-out sets of tens of items make those means coarse, so metric claims are also tested \emph{paired}, exact McNemar over matched per-item verdicts, since evolution rescores the same locked items and two rounds differ only in the grader. That test pools three seeds over one 40-item set, so its unit is the seed-item verdict, not an independent task: it shows the change of grader is not noise, not a task-level population effect. Per-round noise runs near five points, so we report peak and final-rounds values (mean of the last ten) against each run's round-0 baseline. Peak is a locked-set argmax and so optimistic; the final-rounds column is fixed in advance and is the unselected estimate, which is why the paired test is run at both. Bold rows are our proposed configuration.

\subsection{The Evolved Metric Lifts and Transfers}

\begin{figure*}[t]
\centering
\includegraphics[width=\textwidth]{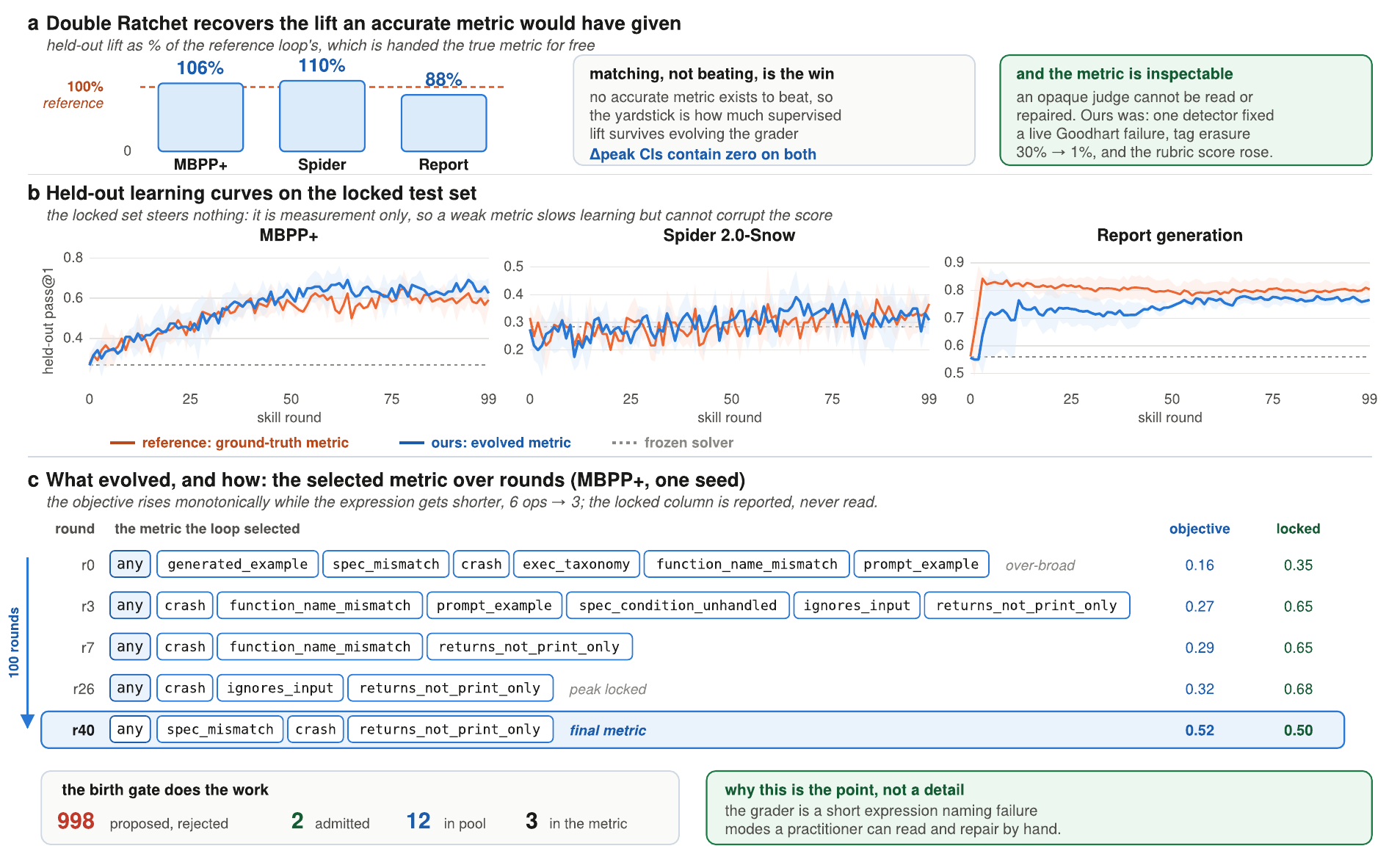}
\caption{\textbf{An evolved metric buys most of the lift ground truth would have, and what evolves stays readable.} \textbf{(a)} Lift Double Ratchet retains as a fraction of the reference loop's; 100\% means the evolved metric bought everything ground truth would have. \textbf{(b)} Locked-set learning curves (mean $\pm$ std, three seeds): reference loop orange, Double Ratchet blue, frozen solver dotted. \textbf{(c)} Every distinct metric the MBPP+ loop selected, in order. The objective rises while the expression shortens from six ops to three; the locked column is reported, never read. Numbers from Tables~\ref{tab:main} and \ref{tab:perloop}.}
\label{fig:results}
\end{figure*}

Figure~\ref{fig:results}c is the mechanism made concrete: every distinct metric the MBPP+ loop selected, in order, as the expression it actually is. The objective rises from 0.16 to 0.52 while the expression \emph{shrinks} from six ops to three, the lifecycle working as designed rather than a search accumulating checks, and the birth gate is the busy part, 998 candidate detectors rejected against 2 admitted on this seed. Table~\ref{tab:oplife} puts that ratio in context across tasks: the pools stay small, the gate does most of the work, and the selected expression is one to three leaves everywhere. What rises monotonically is what selection optimizes: best-so-far selection makes $S(e)$ never fall, on all nine runs (full 100-round budget, early stopping off, last improvement by round 63). Held-out agreement, which selection never sees, is not monotone: five of nine runs end below their own mid-run peak, the ordinary generalization gap of optimizing a ten-item proxy, and why we report the locked column instead of selecting on it. Convergence is always to a composition: \texttt{(any spec\_mismatch crash returns\_not\_print\_only)} on MBPP+ and two-op forms such as \texttt{(any missing\_group\_by spec\_sql\_mismatch)} elsewhere (\hyperref[app:artifacts]{Appendix~E}).

\begin{table}[t]
\centering
\footnotesize
\setlength{\tabcolsep}{2.5pt}
\begin{tabular}{lccccc}
\toprule
Task & Seed & Born & Rejected & Pool & In metric \\
\midrule
MBPP+ & 10 & 2--4 & 997--1014 & 12--14 & 3 \\
Spider & 9 & 24--30 & 612--785 & 33--39 & 1--2 \\
Report & 11 & 11--15 & 172--188 & 22--26 & 2 \\
\bottomrule
\end{tabular}
\caption{\textbf{The birth gate rejects orders of magnitude more detectors than it admits, and selection composes only one to three survivors.} Metric-loop op lifecycle, anchored arm, 100 rounds. \emph{Seed}: hand-authored detectors. \emph{Born}: synthesized and admitted during the run. \emph{Pool}: seed plus born. \emph{In metric}: leaves in the final expression. Three-seed min--max; a bare number means all seeds agree. Skill-side counterpart: Table~\ref{tab:skilllife} in \hyperref[app:lifecycle]{Appendix~F}.}
\label{tab:oplife}
\end{table}

\paragraph{What the lift is measured against.}
These compositions beat the bare LLM judge op they contain, the natural baseline for a grader built without ground truth: peak agreement $0.625\pm0.050$ versus $0.55\pm0.04$ on MBPP+ and $0.500\pm0.026$ versus $0.45\pm0.07$ on Spider (no ground truth exists to score a standalone judge on Report). Round~0 brackets the evolved metric from the other side, being the hand-authored seed composition, at $0.417\pm0.058$ on MBPP+, so the $+0.21$ over it is lift over static engineering rather than lift over nothing. Two baselines would split the remaining credit and neither is run here: a prompt-tuned judge on the same ten items, which would separate anchored composition from anchored prompting, and random composition in place of the LLM composer, which would separate the composer from the search space it draws on.

\paragraph{Three tasks, three roles.}
The tasks are not three attempts at one claim. MBPP+ is the positive case, the only measurable held-out lift; Report starts at ceiling from round 0, where subset seeding already supplies a near-optimal composition, so it carries sufficiency and the safety ablations rather than lift; Spider sits near chance despite a robust objective climb, the detectability boundary taken up in \S\hyperref[sec:detect]{The Detectability Spectrum}. Absolute agreement is not comparable across the three, since each sits on its own locked set with its own reference and class balance, which is why the within-task Spider contrast carries the detectability claim and a cross-task ranking would not.

\paragraph{The lift survives a paired test.}
On MBPP+ the paired test puts the lift beyond seed noise: across 120 matched verdicts, evolution corrects 61 items against 36 broken at peak ($p=0.014$) and 36 against 20 at the end ($p=0.044$), every seed net positive. It finds nothing on Spider, that task's regime rather than a sample-size artifact, since the same test on the same set size detects the MBPP+ effect. Nor is the ten-item anchor fragile: replaying report selection offline on 200 stratified subsamples per size, rescoring every candidate the live run considered, gives held-out agreement $0.871\pm0.063$, $0.882\pm0.051$, and $0.874\pm0.041$ at four, six, and eight anchored items against 0.854 at the full ten. Smaller anchors are noisier but their means do not fall, so recall weighting plus the consensus regularizer keep even a four-item anchor off degenerate metrics. The standard deviations tell the operational story: any single four-item anchor is a lottery, while the mean over draws shows selection is not leaning on one marginal item (protocol in \hyperref[app:negative]{Appendix~G}).

\subsection{Which Guard Carries the Load, and Why Task Score Cannot Tell}
\label{sec:ablation}

Table~\ref{tab:ablation} contrasts three arms of the report metric loop, where the anchor is softest and collusion pressure highest. Disabling the anchor guards (naive) collapses the metric on three of three seeds: selection latches onto a detector that almost never fires, fail-open scoring reports a vacuous perfect objective, and the metric passes 0.97--1.00 of everything it grades. Disabling the lifecycle does not: objectives and held-out agreement stay in the anchored band, and although the pool grows (23--31 ops against 22--26 anchored), the extra ops are inert, since an op never selected never grades anything. This inverts the skill-evolution literature's emphasis: for skills, lifecycle management is the key finding; for evaluators under anchored selection, anchor discipline carries the safety load, so the metric-side analog of library drift is anchor drift, not pool drift. Structurally, a junk skill gets \emph{routed} into prompts and hurts, while a junk op matters only if selection picks it. The lifecycle is not therefore worthless: it keeps the pool and its per-round verdict cost bounded (Table~\ref{tab:oplife}). It is simply not what stands between this system and a worthless grader.

\begin{table}[t]
\centering
\footnotesize
\setlength{\tabcolsep}{2.2pt}
\begin{tabular}{lcccc}
\toprule
Arm & Objective & Held-out & Train & Outcome \\
\midrule
\textbf{anchored} & \textbf{0.865$\pm$0.002} & \textbf{0.830$\pm$0.012} & \textbf{.78--.82} & \textbf{composes} \\
naive & 1.000 (vacuous) & fail-open & .97--1.0 & collapses \\
no-lifecycle & 0.896$\pm$0.072 & 0.868$\pm$0.061 & n/a & no collapse \\
\bottomrule
\end{tabular}
\caption{\textbf{Removing the anchor guards collapses the metric into a vacuous always-pass grader; removing the lifecycle does not.} Metric-validity ablations on report generation, final-rounds values over three seeds. \emph{Objective}: the selection score of Eq.~\ref{eq:score}. \emph{Held-out}: agreement with the locked reference. \emph{Train}: fraction of training attempts the metric passes. The naive arm scores 1.000 because fail-open scoring credits a detector that never fires; it passes essentially everything, so its held-out entry is a behavior rather than a number. Which guard each arm disables is specified in \hyperref[app:lang]{Appendix~C}.}
\label{tab:ablation}
\end{table}

\begin{figure*}[t]
\centering
\includegraphics[width=\textwidth]{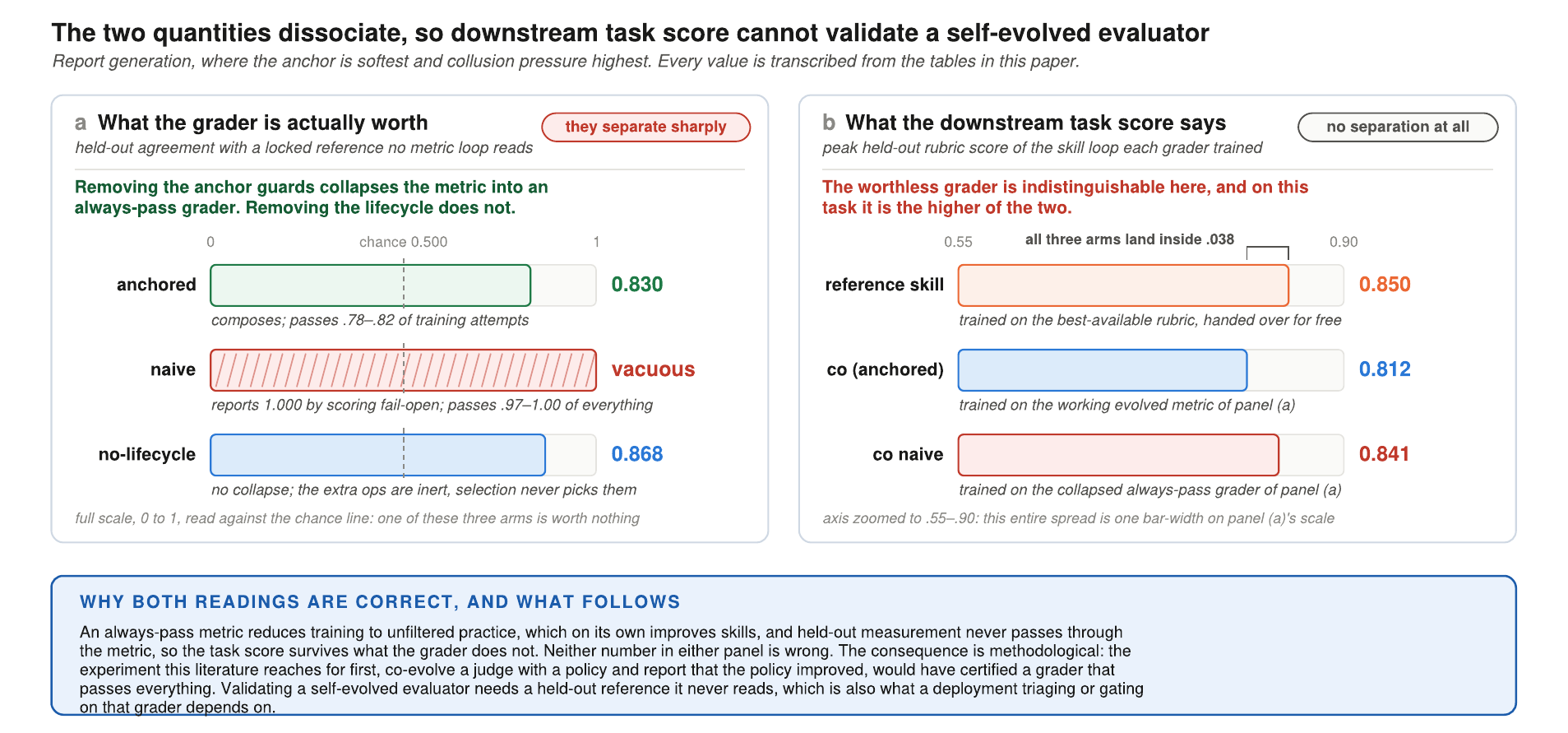}
\caption{\textbf{Metric validity and downstream task score dissociate, so task score cannot certify a self-evolved grader.} The same three report-generation arms, measured two ways. \textbf{(a)} Metric validity on an absolute 0-to-1 scale against a locked reference no metric loop reads: the arms separate sharply, and the naive arm is vacuous rather than merely worse. \textbf{(b)} The task score of the skill loop each grader trained, on a zoomed axis: all three land within .038, and the worthless grader is the higher of the two co-evolved arms. Both readings are correct, which is the problem: certifying a grader by the performance of the policy it trained would have certified the hatched bar. Values from Tables~\ref{tab:ablation} and \ref{tab:main}.}
\label{fig:dissoc}
\end{figure*}

The collapsed arm also carries a warning past this paper, drawn out in Figure~\ref{fig:dissoc}. Its metric is worthless, yet its co-loop scores as well as the anchored one and on two tasks slightly better (Table~\ref{tab:main}). Neither number is wrong: an always-pass metric reduces training to unfiltered practice, which on its own improves skills, and measurement never passes through the metric, so the score survives what the grader does not. The consequence is methodological. \textbf{Downstream task performance cannot validate a self-evolved evaluator}, and the experiment this literature reaches for first, co-evolve a judge with a policy and report that the policy improved, would have certified a grader that passes everything. Validating the evaluator needs a held-out reference it never reads, as above, which is also what a deployment triaging or gating on it depends on.

The asymmetry is worth stating in both directions, because it is easy to over-read. A high task score does not certify the grader, as the naive arm shows. But a low task score does not condemn one either, since a grader can be right and still drive a loop into a capability ceiling it cannot reach, which is what happens on Spider below. Task score and metric validity are simply different measurements of different objects, and the practice of substituting the first for the second is what we are arguing against.

\subsection{Co-Evolution: Is the Evolved Metric Sufficient?}

\begin{table}[t]
\centering
\footnotesize
\setlength{\tabcolsep}{3pt}
\begin{tabular}{lccc}
\toprule
 & MBPP+ & Spider & Report \\
\midrule
Peak: skill & 0.700$\pm$0.025 & 0.483$\pm$0.038 & 0.850$\pm$0.010 \\
Peak: \textbf{co} & \textbf{0.717$\pm$0.038} & \textbf{0.458$\pm$0.038} & \textbf{0.812$\pm$0.006} \\
Peak: co naive$^\dagger$ & 0.742$\pm$0.014 & 0.458$\pm$0.029 & 0.841$\pm$0.003 \\
\midrule
Lift retention (co) & 106\% & 110\% & 88\% \\
$\Delta$peak (co $-$ skill) & $+.02$ & $-.03$ & $-.04$ \\
\quad 95\% CI & $[-.03,.06]$ & $[-.08,.03]$ & $[-.05,-.03]$ \\
\midrule
Improved: skill & 16/23 & 4/12 & 99\% \\
Improved: \textbf{co} & \textbf{19/23} & \textbf{6/12} & \textbf{99\%} \\
\bottomrule
\end{tabular}
\caption{\textbf{Double Ratchet retains 88--110\% of the lift ground truth or a hand-written rubric would have bought.} Held-out task score, three seeds, 40--48-item locked sets. \emph{skill}: the reference loop, trained on ground truth (MBPP+, Spider) or the best-available rubric (Report). \emph{co}: Double Ratchet. \emph{co naive}: the anchor-guards-off ablation, seed-matched to \emph{co}, whose metric is vacuous ($^\dagger$; Table~\ref{tab:ablation}) yet scores as well, which is why this table answers sufficiency but cannot certify a grader (Figure~\ref{fig:dissoc}). \emph{Lift retention}: co's peak lift over round~0 as a fraction of skill's. \emph{$\Delta$peak}: the seed-level difference with a 95\% bootstrap CI, containing zero on both ground-truth tasks. \emph{Improved}: never-passed tasks solved at least once, or (Report) sections ending above round~0.}
\label{tab:main}
\end{table}

\begin{table}[t]
\centering
\footnotesize
\setlength{\tabcolsep}{3pt}
\begin{tabular}{llccc}
\toprule
Task & Loop & Init & Peak & End \\
\midrule
MBPP+ & skill & 0.275$\pm$0.050 & 0.700$\pm$0.025 & 0.592$\pm$0.039 \\
 & \textbf{co} & \textbf{0.267$\pm$0.038} & \textbf{0.717$\pm$0.038} & \textbf{0.652$\pm$0.034} \\
 & co naive & 0.317$\pm$0.058 & 0.742$\pm$0.014 & 0.632$\pm$0.023 \\
\midrule
Spider & skill & 0.317$\pm$0.038 & 0.483$\pm$0.038 & 0.327$\pm$0.010 \\
 & \textbf{co} & \textbf{0.275$\pm$0.043} & \textbf{0.458$\pm$0.038} & \textbf{0.323$\pm$0.038} \\
 & co naive & 0.292$\pm$0.052 & 0.458$\pm$0.029 & 0.343$\pm$0.028 \\
\midrule
Report & skill & 0.562$\pm$0.001 & 0.850$\pm$0.010 & 0.802$\pm$0.024 \\
 & \textbf{co} & \textbf{0.557$\pm$0.006} & \textbf{0.812$\pm$0.006} & \textbf{0.767$\pm$0.015} \\
 & co naive & 0.559$\pm$0.003 & 0.841$\pm$0.003 & 0.811$\pm$0.015 \\
\bottomrule
\end{tabular}
\caption{\textbf{The arms stay ordered at the unselected End column exactly as at peak, so retention is not an artifact of the peak argmax.} Per-loop scores behind Table~\ref{tab:main}, mean $\pm$ std over three seeds. \emph{Init} is round~0, \emph{Peak} the best round on the locked set, \emph{End} the mean of each run's last ten rounds. Frozen single-call baselines are $\approx$0.27 (MBPP+), 0.285 (Spider~2.0), and 0.56 (Report). Bold rows are our proposed configuration.}
\label{tab:perloop}
\end{table}

With validity established independently (\S\hyperref[sec:ablation]{Which Guard Carries the Load}), the remaining question is sufficiency: can a metric evolved from ten reference examples drive a skill loop that ground truth would otherwise drive? It can. Double Ratchet retains 106\%, 110\%, and 88\% of the reference loop's lift on MBPP+, Spider, and Report (Table~\ref{tab:main}; Figure~\ref{fig:results}a--b): within noise on the ground-truth tasks, and a small gap where the reference is itself a \emph{partial rubric}, which is why the independent final judge below carries more weight there. Table~\ref{tab:perloop} gives the underlying init / peak / final-rounds triples, and the ordering of arms at the unselected final-rounds column matches the ordering at peak, so retention is not an artifact of taking an argmax on the locked set.

Spider's absolute lift is smallest for \emph{both} loops in Figure~\ref{fig:results}b, one ceiling binding them alike: a single-call solver's residual failures mostly need live exploration no guidance-text skill expresses. Aggregates also hide a task-level result: both loops solve held-out problems the frozen solver \emph{never} solves, and on both hard-anchor tasks Double Ratchet converts \emph{more} of them than the reference loop (Table~\ref{tab:main}, \emph{Improved}). That settles how good a metric must be to \emph{suffice} here: not very. Spider's metric agrees with ground truth only at $0.500\pm0.026$ yet its co-loop retains the full reference lift, because the metric's training role is \emph{directional}: it decides which attempts become failure capsules, whose concrete error text does not depend on the label being perfect. Metric-graded training accuracy stays in a realistic band (0.78--0.82 on Report, 0.41--0.83 on Spider), so the metric is really grading, and a weak one slows learning without corrupting measurement.

That robustness has a sharp edge, visible on Spider at the seed level: two of three co-evolution seeds selected a weak single-op metric with train agreement 0.41--0.44, yet reached peak evaluations of 0.45--0.50, indistinguishable from the sibling seed carrying a two-op metric (\hyperref[app:negative]{Appendix~G}). A skill loop is a forgiving consumer of grades. The reasons to build a metric well therefore lie elsewhere, in the uses a skill loop does not exercise: triage, where a grader decides which outputs a human sees; gating, where it decides which ship; and repairability, where its structure decides whether a failure can be localized at all. The next subsection puts repairability under load.

\subsection{The Goodhart Episode and the Final-Judge 2$\times$2}
\label{sec:goodhart}

The report task produced the paper's most instructive sequence (Table~\ref{tab:judge}), the clearest evidence that an inspectable metric is worth having. The rubric's metric-discipline dimension counts inline evidence tags, and the first skill runs raised the rubric score by $+0.26$ partly by gaming exactly this: evolved skills wrote the tag \emph{in place of} the number (about 30\% of tags at peak rounds had no value beside them) and invented confident forecasts for a style dimension. The independent judge preferred the baseline in 88\% of decided pairs, citing unrendered placeholders and fabrication. This is a fact about the setup rather than about our system: the rubric was the best available reference signal on this task, hand-written by the pipeline's authors, and gamed within a hundred rounds by skills optimizing against it. Any design that treats such a rubric as ground truth inherits that outcome silently.

The repair was one coupled change: a vocabulary-aware value-erasure check in the capsule gate (a tag whose registered metric has a numeric value but no numeral nearby is a defect) plus the failure hints that teach value-plus-tag, the hint being how the detector's verdict reaches synthesis. We claim the pair, not the detector alone, since separating them needs a detector-only and hints-only rerun we did not do. The rerun cut erased tags to about 1\% while the rubric score \emph{rose} to $0.850\pm0.010$, so nothing was traded away for the fix. Reading the metric is what localized the defect to a nameable check, an intervention a learned scalar does not admit: a reward model that had absorbed the same rubric would have shown a number going up, with no place to put the repair.

The judge itself then needed auditing. The pipeline requires raw evidence tags in its output, a format contract the original generic rubric did not know, so its loss rationales blamed that syntax and the repair barely moved the win rate (0.122 to 0.126). A task-aware rubric stating the contract, still treating valueless tags and fabricated figures as defects, re-judged the same stored pairs: down that column the fix moves the win rate from 0.515 (271 of 526 decided) to 0.770 (435 of 565), real content quality, while across each row the jump measures convention-blindness. That rubric was written after reading the generic judge's rationales, so it diagnoses the audit rather than independently confirming the repair; its pre-repair cell is the control, both cells re-judging identical stored outputs. Both directions matter: self-evolved metrics need an outer audit, and the audit needs the task contract. A deployment that had stopped at the generic column would have concluded the repair did nothing and reverted it.

\begin{table}[t]
\centering
\small
\setlength{\tabcolsep}{4.5pt}
\begin{tabular}{lcc}
\toprule
 & Generic judge & Task-aware judge \\
\midrule
pre-repair (gamed proxy) & 0.122 & 0.515 \\
post-repair (erasure fixed) & 0.126 & \textbf{0.770} \\
\bottomrule
\end{tabular}
\caption{\textbf{Reading down a column shows what the repair bought; reading across a row shows how much the judge penalized a required output format.} Final-judge win rate of evolved report outputs over their pre-evolution baselines, over decided pairs (both position orders agreeing); ties are 39--50\% of the 864 pre-repair and 1{,}056 post-repair pairs. All four cells re-judge the same stored outputs. Both rubrics and the pairing protocol are in \hyperref[app:judge]{Appendix~D}.}
\label{tab:judge}
\end{table}

\subsection{The Detectability Spectrum}
\label{sec:detect}

One variable organizes the results above: how mechanically detectable the solver's failures are. MBPP+ failures are crashes and wrong outputs cheap ops see directly, and reports have a closed failure taxonomy behind their anchor (\hyperref[app:setup]{Appendix~B}), so transfer is strong on both. Spider~2.0 turns that into an in-task controlled contrast. Under our first prompt most failures were compile errors and the evolved metric reached 0.85 held-out agreement, but grounding the prompt with typed schemas raised the solver's baseline by half and left \emph{residual} failures that are semantically wrong values under clean execution, invisible to deterministic ops, so agreement fell to $0.500\pm0.026$ even as the objective lifted. Same method, same task, different failure distribution: a semantic regime shifts the burden onto judge ops and outer audits, and the sharper the solver, the more of this regime is left. What evolves stays legible either way, which is what made the repair possible.

\begin{figure*}[t]
\centering
\includegraphics[width=\textwidth]{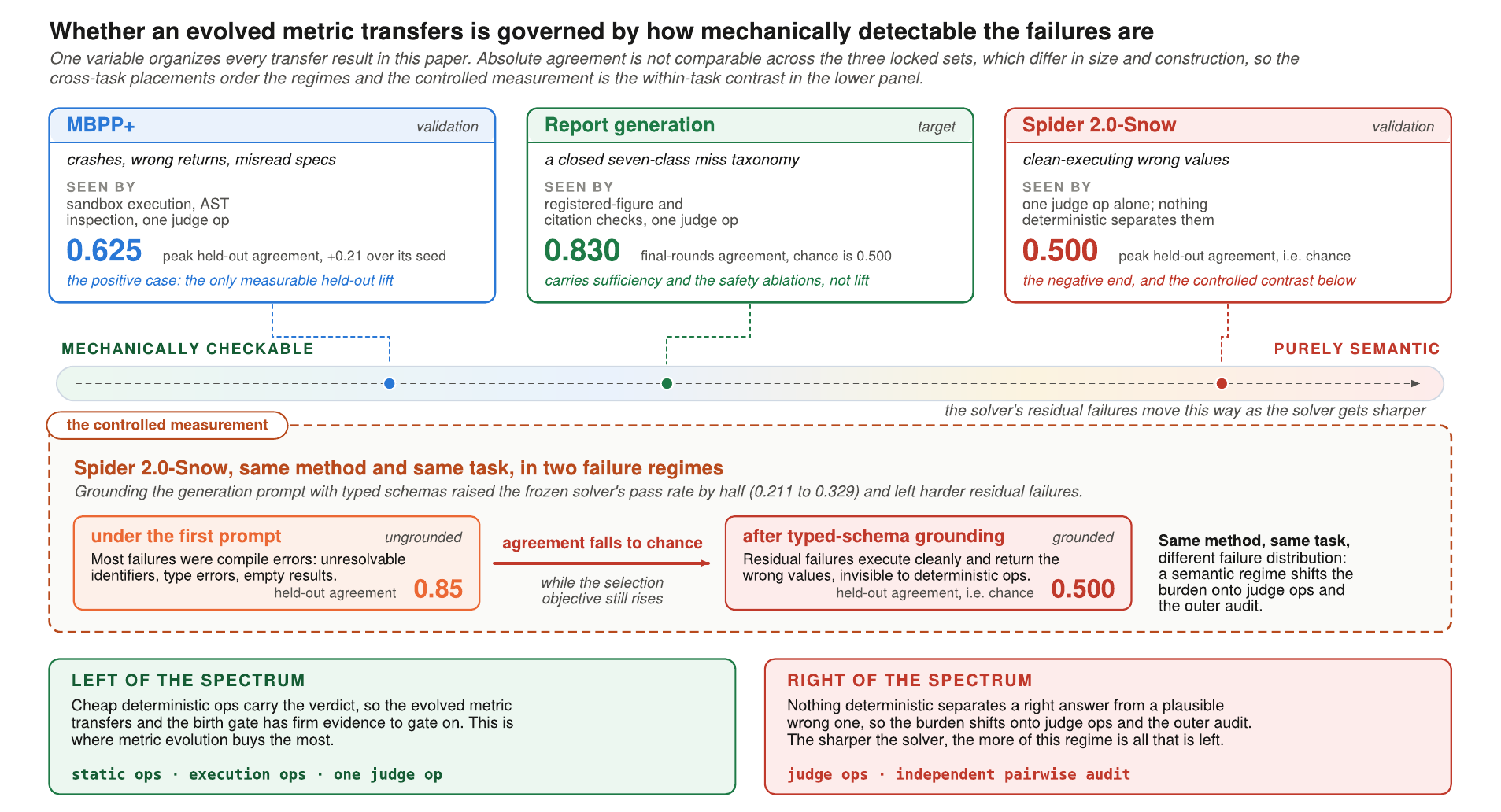}
\caption{\textbf{Whether an evolved metric transfers is governed by how mechanically detectable the solver's failures are.} \emph{Top:} the three task families on the axis from mechanically checkable failures to purely semantic ones, each with its evolved metric's peak held-out agreement. Those three numbers sit on three different references, so the placement is an interpretation, not a measurement. \emph{Middle, the measurement:} a within-task Spider~2.0-Snow contrast. Grounding the solver prompt with typed schemas pushes the residual failures from compile errors into clean-executing wrong values, and the same metric loop's held-out agreement falls from 0.85 to 0.500. \emph{Bottom:} what each end of the axis implies for practice.}
\label{fig:detect}
\end{figure*}

Figure~\ref{fig:detect} lays the three task families on that axis and keeps the controlled comparison separate from the cross-task placement, because only the former is a measurement. The placement of MBPP+, Report, and Spider on the spectrum is an interpretation of where each family's failures sit; their agreement numbers are not comparable to one another, since each is computed against a different locked reference on a different set. The Spider band underneath is the actual evidence: one task, one method, two solver prompts, and the metric's held-out agreement moving from 0.85 to 0.500 as the residual failures move from mechanical to semantic.

The practical reading is a scoping rule. Where failures are mechanically checkable, cheap deterministic detectors carry the metric and an evolved composition transfers; where they are purely semantic, the composition still evolves and still names what it finds, but its agreement with truth is bounded by the judge ops inside it, and the outer audit stops being optional. The uncomfortable corollary is that this boundary moves in the direction of harder, not easier: every improvement to the solver retires the failures a static op could catch and leaves behind the ones only judgment can.

\section{Discussion}

\paragraph{The anchor cannot be manufactured.}
Evolution expands coverage but never creates ground truth. With no anchor the loop optimizes noise, which is what the naive arm demonstrates from the inside: remove the guards that tie selection to ten real labels and the search finds the cheapest way to look perfect. Anchor \emph{quality} therefore matters as much as its existence, and hardening soft anchors into checkable ones is the highest-leverage investment available. Ten items is few; ten items with golden references and a strict soft-labeling protocol were enough on all three families, and the subsample replay says four would often have been.

\paragraph{Validity and sufficiency are different questions.}
The two axes of the results section are not two views of one number. Sufficiency asks whether a loop trained on the evolved metric reaches where ground truth would take it, and the answer is largely yes, because a skill loop consumes grades directionally. Validity asks whether the grader is right, and it is the question every use outside a training loop turns on. Conflating them is not a philosophical error but a measurable one: Figure~\ref{fig:dissoc} shows a grader that answers the first well and the second not at all.

\paragraph{Where the method helps.}
Metric evolution buys the most where the solver's failures are mechanically detectable, as the within-task Spider contrast quantifies (\S\hyperref[sec:detect]{The Detectability Spectrum}). It buys least where they are semantic, and there its value shifts from agreement to legibility: an expression that names its verdicts supports triage and repair even when it barely beats chance at predicting truth.

\section{Limitations}

This is a mechanism study, not a scaling result. Three task families spanning the detectability axis including its negative end, one model family, rounds in the low hundreds, and held-out sets of tens of items, though graded every round by every candidate (52{,}600 item-level verdicts against the locked reference and 5{,}743 birth-gate decisions). Nothing in the design is model-specific, and every LLM role is bound from a configuration table, but we have not run a cross-family replication and cannot claim one.

Several specific gaps are worth naming. Teacher quality is held fixed, so how anchor quality degrades with a weaker teacher is untested. The paired McNemar test pools seeds over a single 40-item locked set, so it licenses a claim about graders, not about tasks. Peak values are locked-set argmaxes and therefore optimistic, which is why the final-rounds column, fixed in advance, is reported beside them everywhere. The Goodhart repair is claimed as a coupled pair of detector and hints, since we did not run the detector-only and hints-only arms that would separate them. The task-aware rubric was written after reading the generic judge's rationales, so it is a diagnosis and not an independent confirmation. Guidance-text skills cannot substitute for capabilities that need interaction, such as live schema exploration, which is the ceiling Spider's absolute lift runs into for both loops. And metamorphic detectors \citep{chen1998metamorphic}, the natural next op family for exactly the semantic regime where our current pool is weakest, are built but unexercised here.

\section{Conclusion}

The evaluation metric can itself be the evolving object, and it evolves into something valid: $+0.21$ held-out agreement on MBPP+ against a locked reference no part of selection reads, ahead of the bare judge it contains, and beyond seed noise under a paired test. Anchor discipline, not the pool lifecycle, makes that hold, inverting the lesson the skill-evolution literature would predict, and the arm without it warns that task score certifies nothing, a worthless metric training skills just as well. Read for sufficiency rather than validity, Double Ratchet retains 88--110\% of the supervised lift across code, enterprise SQL, and report writing. On the one task where no golden metric exists, an outer judge caught the loop gaming its own reference, one detector plus its hints repaired it without cost to the score, and auditing the judge caught the audit mistaking a required output format for a defect. Who grades the grader: an anchor it predicts but never sees, and an outside judge that must itself be checked.

\clearpage
\appendix

\section{Appendix A: Task Descriptions}
\label{app:tasks}

These cards give enough context to interpret the results; the builders that assemble each family's tasks and splits are in the released code.

\begin{taskbox}{MBPP+\cardsub{validation task $\cdot$ anchor: hidden unit tests}}
\small
\textbf{Task.} A one-paragraph natural-language programming problem plus one example call; the solver writes a Python function.\\[2pt]
\textbf{Anchor.} The EvalPlus hidden test suite, which adds many more tests per problem than the original MBPP and repairs faulty ones, giving reliable pass/fail.\\[2pt]
\textbf{Typical failures.} Crashes, mishandled edge cases, printing instead of returning, misread specifications, exactly the failure classes the op pool checks.
\end{taskbox}

\begin{taskbox}{Spider 2.0-Snow\cardsub{validation task $\cdot$ anchor: execution match}}
\small
\textbf{Task.} An analyst question over a real hosted Snowflake warehouse: production-scale databases with hundreds of tables, cryptic columns, sharded table families, and dialect-specific SQL. The solver writes one query from the question, a rendered schema catalog, and optional domain notes.\\[2pt]
\textbf{Anchor.} Execute the candidate and compare the fetched table to the stored gold result with the official comparator (value-based; column names and order are ignored).\\[2pt]
\textbf{Typical failures.} Unresolvable identifiers, type errors on date-like columns, and, hardest, queries that run cleanly but compute the wrong thing. Our single-call solver passes about a third of screened tasks.
\end{taskbox}

\begin{taskbox}{Report generation\cardsub{target task $\cdot$ no accurate metric exists}}
\small
\textbf{Task.} Write one \emph{section} of an analyst report (e.g., ``value-chain map: control points from ASIC to optical engine'') in markdown, given the inputs below.\\[2pt]
\textbf{Inputs.} (i)~Section title and purpose. (ii)~A style contract fixing units, reference dates, and accounting rules (e.g., ``all amounts USD, calibrated to 2026~Q2, market sizes must not be summed across pools''). (iii)~Evidence cards. (iv)~A metric vocabulary registering the section's key figures (id, name, value, unit).\\[2pt]
\textbf{Evidence card.} An audited unit of evidence: id \texttt{E0103}, one-sentence claim, verbatim source quote, source attribution and date, stance (supporting / opposing / neutral), and an auditor's credibility verdict.\\[2pt]
\textbf{Requirements.} Ground every claim in the cards (cite \texttt{[E\#\#\#]}); state each key figure's value with its inline tag binding it to the metric vocabulary (rendered downstream as audit tooltips); hedge weak or disputed evidence and present both sides of conflicts; include required structure such as a summary line.\\[2pt]
\textbf{Why no metric exists.} Section quality is holistic; even an expert author cannot reduce ``a good section'' to a checkable rule. The RAQS rubric scores lexical and structural proxies of the requirements above and is treated as reference signal only.
\end{taskbox}

\section{Appendix B: Setup Details}
\label{app:setup}

\paragraph{Splits and screening.}
For MBPP+ and Spider~2.0 we screen every task with five independent samples of the frozen single-call solver, graded by the official anchor. We drop tasks the solver always passes (no room to improve) and stratify the rest across pass levels for a balanced difficulty mix, then split 60/40 into train and test. The dev set is ten items, five clear failures and five clean passes (for report generation: five golden-quality positives and five class-stratified checkable failures), so the metric is scored on both classes. In the released code the dev and test splits are named \texttt{eval\_dev} and \texttt{eval\_locked}.

\paragraph{Anchors.}
\emph{MBPP+}: hidden unit tests. \emph{Spider~2.0}: execute the candidate on the live warehouse and compare the fetched table to the stored gold result with the official comparator, which matches values (ignoring column names and order). We reproduce the evaluator's CSV round-trip of the fetched frame, without which even gold queries fail on dtype. \emph{Report generation}: the metric loop's anchor is a frozen demo-delta soft label, where a teacher compares the frozen-agent report against the golden section under a closed seven-class miss taxonomy and passes iff at most one miss class is present. Each of the 48 held-out sections contributes two locked items, the frozen-agent report (which the strict anchor almost always fails) and the golden section (which passes by definition), giving 96 items balanced 50/50, so a degenerate metric scores 0.5. The skill and co-loops are evaluated by the official rubric evaluator (RAQS, 0--100 normalized), whose dimensions cover evidence grounding, metric-tag discipline, claim calibration, structure, and style.

\paragraph{Spider~2.0 prompt grounding.}
The generation prompt grounds the solver with a compact typed schema catalog (columns as name:TYPE, sharded tables collapsed, sample values on date-like columns) and Snowflake-specific rules (quoted identifiers, type-respecting casts, value-only grading). We report this because it is a large, shared effect we control for, not a contribution: grounding raises the frozen baseline by half on the screened pool (0.211 to 0.329) but is applied identically to every arm, so all reported lifts sit on top of it. It also sets up the detectability contrast: grounding pushes the solver's residual failures from compile errors into clean-executing wrong values.

\section{Appendix C: The Metric Language and Lifecycle}
\label{app:lang}

\paragraph{Language.} Recapping the grammar of Eq.~\ref{eq:grammar}: a metric is an expression over ops (pure functions of (task, output) returning \textsc{drawback}/\textsc{clean}/\textsc{abstain}), composed by $\bigvee$, $\bigwedge$, negation, the unweighted $K$-of-$k$ vote, and a weighted vote against a threshold, with abstaining children excluded from a combinator and an all-abstain node abstaining. Negation and the weighted vote are offered to the composer on every round but appear in none of the selected metrics, so every reported expression reads as a plain drawback disjunction or vote.

\paragraph{Lifecycle.} Each task starts from a small pool of seed detectors, the obvious failure checks a practitioner reads straight off the task description (does the code crash, does the query group correctly, is a key figure missing); drafting them, or having an LLM draft them from the task spec, is an afternoon's work, and their exact contents barely matter because the lifecycle grows and prunes the pool from there. Synthesis authors typed op specs from clustered misses under a closed per-task taxonomy. The birth gate requires a new op to fire on at least half its cluster and stay clean on known-good outputs. Ops born from anchored misses enter active; unlabeled-gap ops enter shadow and are promoted only on demonstrated anchor agreement. Per-op fitness is the leave-one-out marginal of the incumbent expression, and non-positive ops retire after a two-round grace.

\paragraph{Selection and guards.} The objective multiplies dev agreement by consensus agreement on unlabeled train outputs (exponent 1.0 for MBPP+ and Spider, 0.25 for report generation, where op consensus is a weaker regularizer), minus a small complexity penalty. Both agreement terms are weighted. Dev agreement is the recall-weighted mean of the two per-class recalls over the dev items a candidate opines on, with the \textsc{fail} class at weight 2 and the \textsc{clean} class at weight 1. Train consensus is reliability-weighted: each opining op's vote is scaled by $1 + 10 \max(0, m_o)$ for that op's anchored leave-one-out marginal $m_o$, so anchor-proven ops lead the vote and ops with no anchored evidence contribute at unit weight. This keeps a bloc of cheaply synthesized, correlated detectors from defining the pseudo-label that the regularizer rewards agreeing with. The complexity penalty charges 0.001 per op beyond a floor of three and a further 0.002 per LLM-calling op, which is why selected expressions stay small and mostly deterministic. Fail-closed anchoring makes candidates without a usable dev opinion unselectable, and a validity gate drops all-pass/all-fail/all-abstain candidates. The naive ablation disables exactly these two guards; the no-lifecycle ablation disables exactly the birth gate, shadow tier, and retirement.

\section{Appendix D: Final-Judge Protocol and Rubrics}
\label{app:judge}

\paragraph{Protocol.} For each of the 48 held-out sections we pair the stored round-0 report against the stored peak-round report, and separately against the final-round report, for every arm and seed. Pre-repair this is 3 arms $\times$ 3 seeds $\times$ 2 rounds $\times$ 48 sections $=$ 864 pairs; the post-repair rerun adds a fourth seed on the two co-loop arms, giving 1{,}056. Each pair is judged twice with positions swapped, and a win counts only when both orders agree; otherwise it is a tie.

\paragraph{Rubrics.} The generic rubric asks a senior-editor question about holistic reader value with no knowledge of the pipeline's conventions. The task-aware rubric additionally states the format contract: evidence citations and inline metric tags are required output that renders downstream; a tag with no value nearby is a defect; fabricated figures are disqualifying. Both rubrics judge the same stored pairs, so the four cells of Table~\ref{tab:judge} differ only in the repair pass (rows) and the rubric (columns).

\section{Appendix E: What Evolution Actually Produced}
\label{app:artifacts}

Final artifacts from the paper runs, verbatim except where marked as condensed, illustrating the central qualitative claim: everything the loops evolve is legible. Figure~\ref{fig:evolved} gives the final metric on each task, Figure~\ref{fig:oplife} traces one synthesized detector through the whole lifecycle, Figure~\ref{fig:grading} grades one attempt with the hidden tests and with the evolved metric side by side, and Figure~\ref{fig:samelesson} pairs skills the two loops learned independently.

\begin{figure*}[t]
\begin{taskbox}{Evolved metrics\cardsub{one final expression per task{,} verbatim}}
\small
\setlength{\tabcolsep}{0pt}
\renewcommand{\arraystretch}{1.15}
\begin{tabular}{@{}>{\raggedright\arraybackslash}p{0.30\linewidth}@{\hspace{14pt}}>{\raggedright\arraybackslash}p{0.29\linewidth}@{\hspace{14pt}}>{\raggedright\arraybackslash}p{0.34\linewidth}@{}}
\textbf{MBPP+} & \textbf{Spider 2.0} & \textbf{Report} \\
\addlinespace[2pt]
Flag a solution if the judge finds a spec mismatch, or it crashes on probe inputs, or it prints instead of returning:
&
Flag a query if it aggregates without grouping, or the judge finds it does not answer the question (selected independently on two of three seeds):
&
Flag a section if a registered key figure never appears, or a weak-credibility card is cited without hedging (the other two seeds swap the first leaf for an uncited-claims detector):
\\
\addlinespace[4pt]
\texttt{(any}\par
\texttt{~~spec\_mismatch}\par
\texttt{~~crash}\par
\texttt{~~returns\_not\_print\_only)}
&
\texttt{(any}\par
\texttt{~~missing\_group\_by}\par
\texttt{~~spec\_sql\_mismatch)}
&
\texttt{(any}\par
\texttt{~~ledger\_metric\_uncovered}\par
\texttt{~~low\_status\_card\_overclaimed)}
\\
\end{tabular}
\end{taskbox}
\caption{\textbf{The final evolved metric on each task, verbatim, with a plain reading above it.} Each is a short drawback disjunction over typed detectors: nothing here is a learned scalar, and each leaf names the failure class it fires on.}
\label{fig:evolved}
\end{figure*}

\begin{figure*}[t]
\begin{taskbox}{A synthesized op\cardsub{born{,} gated{,} promoted{,} and selected (Spider 2.0)}}
\small\raggedright
Tracing one op through the whole lifecycle, on the metric-loop seed that ends at \texttt{(any syn\_forbids\_select\_star\_when\_columns spec\_sql\_mismatch)}. The loop clustered four unlabeled train outputs that returned every column although the question implied specific ones, and authored the typed spec \texttt{\{kind: static, check: forbids\_select\_star, detects: select\_star\_specific\_columns\}}. Because its birth evidence was an unlabeled gap rather than an anchored miss, it entered as a \emph{shadow}: it ran and recorded verdicts but could not be selected. It was promoted the same round only once composing it with the incumbent raised dev agreement from 0.71 to 0.81, and selection then adopted it, replacing the grouping check the seed pool had contributed. The two co-loop seeds on this task converge on a different synthesized op (\texttt{syn\_top\_n\_missing\_limit}) and the third on the seed \texttt{select\_star} check, so the leaf is not a fixed artifact of the pool.
\end{taskbox}
\caption{\textbf{One synthesized detector through the whole lifecycle: born from unlabeled outputs, admitted as a shadow, promoted only once it raised anchor agreement, then selected.} A Spider~2.0 metric-loop seed; the aggregate counts are Table~\ref{tab:oplife}.}
\label{fig:oplife}
\end{figure*}

\begin{figure*}[t]
\begin{taskbox}{Grading one task: hidden tests vs.\ evolved metric\cardsub{MBPP+ task{,} final-round trace{,} verbatim}}
\small
\setlength{\tabcolsep}{0pt}
\renewcommand{\arraystretch}{1.15}
\begin{tabular}{@{}>{\raggedright\arraybackslash}p{0.36\linewidth}@{\hspace{14pt}}>{\raggedright\arraybackslash}p{0.27\linewidth}@{\hspace{14pt}}>{\raggedright\arraybackslash}p{0.31\linewidth}@{}}
\textbf{Task and solver output} & \textbf{Ground-truth anchor} & \textbf{Evolved metric} \\
\addlinespace[2pt]
``Write a function to find the ratio of zeroes in an array of integers.''
\par\vspace{4pt}
\texttt{def zero\_count(nums):}\par
\texttt{~~zeros = sum(1 for x}\par
\texttt{~~~~~~~~~~in nums if x == 0)}\par
\texttt{~~non\_zeros = len(nums) - zeros}\par
\texttt{~~...}\par
\texttt{~~return zeros / non\_zeros}
\par\vspace{4pt}
The code runs cleanly and returns a number; it computes the ratio to \emph{non-zeros} where the spec wants the ratio to \emph{all} elements. Only the meaning is wrong.
&
Hidden unit tests (EvalPlus).
\par\vspace{4pt}
\texttt{assert zero\_count(...)}\par
\texttt{~~$\rightarrow$ AssertionError}
\par\vspace{4pt}
Verdict: \textsc{fail}.
\par\vspace{4pt}
The verdict is reliable but opaque: no indication of which requirement was misread.
&
Evolved composition \texttt{(any spec\_mismatch crash returns\_not\_print\_only)}.
\par\vspace{4pt}
\texttt{crash}: clean\par
\texttt{returns\_not\_print\_only}: clean\par
\texttt{spec\_mismatch} (judge op): \textsc{drawback}
\par\vspace{4pt}
Verdict: \textsc{fail}, agreeing with the anchor, and \emph{naming} the failure class: the specification was misread.
\\
\end{tabular}
\par\vspace{5pt}
The named drawback is what turns this attempt into a usable failure capsule for skill synthesis; the anchor's bare verdict could not have driven the same repair.
\end{taskbox}
\caption{\textbf{The same verdict, reached two ways: the hidden tests say \textsc{fail}, and the evolved metric says \textsc{fail} while naming why.} One MBPP+ attempt whose code runs cleanly and returns a number but computes the wrong quantity. Only the middle column is ground truth; the right column is what an inspectable composition adds on top of an equally correct verdict.}
\label{fig:grading}
\end{figure*}

\begin{figure*}[t]
\begin{taskbox}{Same lesson{,} two loops\cardsub{functionally matched skills{,} reference vs.\ co-evolved{,} condensed}}
\small
Each pair below was synthesized independently, one by the reference loop under ground-truth grading and one under the evolved metric (Double Ratchet); neither loop saw the other. Converging on the same task lesson is the qualitative face of the retention result.
\par\vspace{5pt}
\setlength{\tabcolsep}{0pt}
\renewcommand{\arraystretch}{1.15}
\begin{tabular}{@{}>{\raggedright\arraybackslash}p{0.31\linewidth}@{\hspace{14pt}}>{\raggedright\arraybackslash}p{0.31\linewidth}@{\hspace{14pt}}>{\raggedright\arraybackslash}p{0.31\linewidth}@{}}
\textbf{MBPP+} & \textbf{Spider 2.0} & \textbf{Report} \\
\emph{(grader imports a literal identifier)} & \emph{(empty result diagnosis)} & \emph{(the anti-erasure lesson)} \\
\addlinespace[4pt]
\emph{reference} \ \texttt{function-name-} \texttt{from-assert}: ``the harness imports the symbol literally as written in the assert; lift the identifier verbatim including case, the prose is only a hint.''
&
\emph{reference} \ \texttt{sql-empty-result-} \texttt{diagnosis}: ``a 0-row result is almost never correct when the gold has rows; incrementally relax predicates to locate which condition eliminates all rows.''
&
\emph{reference} \ \texttt{metric-tag-annotates-} \texttt{not-replaces}: ``the scorer counts a figure only when a literal number sits adjacent to the brace tag; a tag with no rendered numeral reads as an unrendered placeholder.''
\\
\addlinespace[6pt]
\emph{co} \ \texttt{copy\_callable\_} \texttt{identifier\_from\_assert}: ``the grader imports the name verbatim from the assert; reproduce that exact spelling letter for letter, including apparent typos.''
&
\emph{co} \ \texttt{sql-empty-result-} \texttt{sanity-probe}: ``an empty result almost always means one predicate matched nothing; validate each CTE's row count in isolation before trusting the join.''
&
\emph{co} \ \texttt{metric-tag-inline-} \texttt{after-number}: ``the tag annotates the value; it never replaces the value, and a bare card cite does not satisfy it.''
\\
\end{tabular}
\end{taskbox}
\caption{\textbf{Trained on ground truth and trained on the evolved metric, the two loops independently learn the same task lesson.} One functionally matched skill pair per task, condensed. Neither loop saw the other, so the convergence is the qualitative face of the lift-retention result in Table~\ref{tab:main}.}
\label{fig:samelesson}
\end{figure*}

\section{Appendix F: Lifecycle Statistics}
\label{app:lifecycle}

Table~\ref{tab:oplife} in the results gives the metric-loop op lifecycle; Table~\ref{tab:skilllife} here gives the skill-loop side, both for the anchored arms. The locked metric sets are small, 40 items on MBPP+ and Spider and 96 on reports, but they are regraded in full every round: the nine anchored metric-loop runs record 52{,}600 item-level verdicts against the locked reference and 5{,}743 birth-gate decisions. Each run stores every verdict with its per-op trace, which is what makes the paired tests recomputable from a finished run without rerunning any loop; the released code includes the script that reads those tables and computes the reported counts and $p$-values. For the skill loops, synthesis is prolific (roughly 1.5 skills per round), the bounded bank forces contribution-based eviction of more than half of everything created, and rollback fires rarely; the report skill loop is the outlier, its targeted failure capsules producing skills so effective that a handful suffice (one seed held its plateau with a bank of six).

\begin{table}[t]
\centering
\footnotesize
\setlength{\tabcolsep}{2.2pt}
\begin{tabular}{llcccc}
\toprule
Task & Loop & Synthesized & Bank & Evicted & Rollback \\
\midrule
MBPP+ & skill & 140--167 & 50 & 80--100 & 1 \\
 & \textbf{co} & \textbf{111--154} & \textbf{50} & \textbf{60--101} & \textbf{0} \\
Spider & skill & 149--177 & 50 & 86--114 & 1--2 \\
 & \textbf{co} & \textbf{152--153} & \textbf{50} & \textbf{76--86} & \textbf{1--2} \\
Report & skill & 6--66 & 6--50 & 0--15 & 0 \\
 & \textbf{co} & \textbf{135--156} & \textbf{50} & \textbf{85--106} & \textbf{0} \\
\bottomrule
\end{tabular}
\caption{\textbf{Synthesis is prolific and the bounded bank evicts more than half of what it creates.} Skill-loop lifecycle over 100 rounds, bank cap 50, the counterpart to Table~\ref{tab:oplife}. \emph{Bank}: final active skills. \emph{Evicted}: removed by contribution. \emph{Rollback}: held-out regressions reverted. Three-seed min--max; a bare number means all seeds agree. Bold rows are our proposed configuration.}
\label{tab:skilllife}
\end{table}

\section{Appendix G: Negative and Supporting Results}
\label{app:negative}

\paragraph{Naive co-loops are not a free pass.}
Although naive co-loops match the anchored arm on held-out evaluation (their measurement anchor is untouched), their metric arms are degenerate where inspection matters: on report generation the naive metric grades 0.97--1.00 of train as passing (versus 0.78--0.82 anchored), and on Spider~2.0 the naive objective barely moves on two of three seeds.

\paragraph{Skill-loop robustness to metric quality.}
The weak-metric Spider~2.0 seeds reported in the results reach the same peak as their two-op sibling because training signal quality degrades gracefully: a failed attempt still carries concrete error text into skill synthesis even when the pass/fail label is noisy.

\paragraph{Anchor size.}
Ten dev items is a design choice about what sparse anchoring realistically supplies, not a tuned setting, so we test how far selection depends on it. The results section reports the numbers; the protocol behind them is an offline replay of report-generation selection, drawing 200 stratified subsamples per anchor size from the same ten items, preserving the fail/clean balance within each draw, and rescoring every candidate expression the live run actually considered under the reduced anchor. No new generation and no new solver calls are involved, so the comparison isolates the anchor's contribution to selection from run-to-run variance.

\section{Appendix H: Reproducibility}
\label{app:repro}

The code is released at \url{https://github.com/amazon-science/Self-Evolving-Agents-Double-Ratchet}: all loops, op packs, split builders, screening scripts, and the judge protocols, sufficient to rerun every experiment end to end and reproduce results in line with those reported here. We release code only, not run data, results, or logs. All model calls go through the Amazon Bedrock API, so wall-clock time is dominated by API latency. Each run writes its own round-by-round history, op pool, skill bank, and judge outputs as structured result files, and every number in this paper is read directly from those files. The role-to-model table named in the setup covers every LLM role: solver, teacher, composer, skill synthesizer, judge ops, and final judge. Spider~2.0 requires credentialed access to the benchmark's hosted warehouse; all other experiments are self-contained.

\end{document}